\documentclass[conference]{IEEEtran}

\IEEEoverridecommandlockouts

\usepackage{url}
\usepackage[colorlinks=true, urlcolor=blue, linkcolor=black, citecolor=black]{hyperref}
\usepackage{cite}
\usepackage{amsmath,amssymb,amsfonts}
\usepackage{multirow}
\usepackage{algorithmic}
\usepackage{graphicx}
\usepackage{textcomp}
\usepackage{xcolor}
\usepackage{balance}
\usepackage{booktabs}
\usepackage{fancyhdr}
\usepackage{orcidlink}

\def\BibTeX{{\rm B\kern-.05em{\sc i\kern-.025em b}\kern-.08em
    T\kern-.1667em\lower.7ex\hbox{E}\kern-.125emX}}

\begin{document}


\fancypagestyle{firstpage}{
\fancyhf{}

\fancyhead[L]{\scriptsize
2026 IEEE International Conference on Signal Processing, Information, Communication and Systems (SPICSCON)\\
13-14 August 2026, Faculty of ECE, Bangladesh Army University of Engineering \& Technology (BAUET), Qadirabad, Natore-6431, Bangladesh
}

\fancyfoot[L]{\scriptsize
979-8-3195-1270-3/26/\$31.00 \copyright 2026 IEEE
}

\renewcommand{\headrulewidth}{0pt}
\renewcommand{\footrulewidth}{0pt}

}


\title{Ideological Stance Detection in a Low-Resource Language: Polarization in Bangladeshi Public vs Private University Discourse on Social Media}


\author{

\IEEEauthorblockN{
Safaruzzaman Shovo\textsuperscript{1}\orcidlink{0009-0006-2321-779X},
Monowar Islam\textsuperscript{1}\orcidlink{0009-0003-8376-0012},
Asif Hossain\textsuperscript{1},
Sameya Akhter\textsuperscript{1},
Md. Shamsul Islam\textsuperscript{1}
}

\IEEEauthorblockA{
\textsuperscript{1}\textit{Department of Computer Science and Engineering, Faridpur Engineering College, 
Faridpur, Bangladesh}\\
Email: shovo.research@gmail.com, dew.shishir5000@gmail.com, asifhossain.cs@gmail.com,\\
samiacse08@gmail.com, shamsul.cse@sec.ac.bd
}

}

\maketitle

\thispagestyle{firstpage}

\vspace{-0.3cm}

\begin{abstract}

Public vs. private universities is a debatable issue, and it creates polarization on social media in Bangladesh. Debate on quality, jobs, and prestige is passionate among the students, parents, and graduates, the majority of whom speak Bengali, a low-resource language. To measure this polarization, this paper introduces a manually annotated dataset of 4,060 Bengali comments labeled as Pro-Public, Pro-Private, or Neutral. We evaluated the quality of our annotations by Fleiss's Kappa agreement that was 0.89, corresponding to a high agreement among annotators. The classical ML (SVM, Random Forest, XGBoost), BiLSTM network, hybrid BanglaBERT+XGBoost models and the state-of-the-art zero-shot LLMs (Claude Sonnet 4, DeepSeek-V3.1, Llama 4 Maverick, Kimi K2 Thinking, Qwen3-235B Thinking) models are evaluated. The accuracy of BanglaBERT+XGBoost is 91.81\% and macro F1 score is 91.70\%, which is higher than all the supervised baselines. The zero-shot Llama 4 Maverick Thinking achieves a macro F1 of 0.931 (overall accuracy of 93.31\%) without any fine-tuning. All machine learning (ML), deep machine learning (DL) and transformer models were outperformed by the zero-shot Llama 4 Maverick model. Polarization also is evident, in some ways more clearly in the Pro‑Private comments, which emphasize modern facilities and timely graduation, versus the Pro‑Public comments, which emphasize affordability and government jobs. Our findings open new directions for analyzing social media polarization in low‑resource languages.

\end{abstract}
\renewcommand\IEEEkeywordsname{Keywords}

\begin{IEEEkeywords}
Low-resource NLP, Ideological Stance Detection, Bengali Language, Bangladeshi Educational Discourse, Public vs Private Universities, Social Media Polarization Analysis
\end{IEEEkeywords}

\vspace{-0.2cm}
\section{Introduction}
The debate over whether public or private universities provide better education remains active in Bangladesh. Rather, a few places are polarized with this polarization. Public universities like University of Dhaka, Bangladesh University of Engineering and Technology (BUET), University of Chittagong, and Rajshahi University have been considered as cradles of meritocracy for generations, churning out professionals, academics, and civil servants. These institutions have low tuition costs, have the highest performing students on standardized exams, and have institutional prestige spanning decades. They also face challenges due to delays in sessions, conflicts, lack of seats, and out-dated infrastructure.

However, in the 1990's some private universities (such as North South University (NSU), BRAC University, Independent University Bangladesh (IUB), East West University) came into existence and have grown rapidly. They provide the latest facilities, English classes, and timely graduation, and they claim that their courses focus on industry needs. But they are criticized for charging too much tuition, the quality of the institutions and that they take in students who missed out on public university spots.

This tension is always at play, in social media. It seems that almost every time a student posts a letter of admission, a person mentions a job prospects or a news report surfaces on university rankings, the comment section will be filled with debate on which system is better at producing better graduates. It can get very personal, as advocates of public universities say private university students are buying degrees and those of private universities scoff on public university students as unemployable rote learners.

There are many studies that have looked at educational preferences through traditional survey techniques\cite{amin2021comparative}\cite{biswas2026adequacy} but none of them have focused on organic conversation. Social media analysis also sheds light on people's actual arguments on these subjects when discussing them with each other, not with the researcher who has asked a question. Though some studies have worked on ideological stance detection in political contexts related to polarization\cite{ali2026social}, very little work has been done on educational polarization, especially in non-Western contexts.

To fill the void, we do a through analysis of Bengali comments, where people talk about public and private universities. The ideological stance of 4,060 comments is manually annotated and with high inter-annotator reliability. Then, we compare the performance of a number of machine learning techniques, including neural networks, pre-trained language models fine-tuned for Bengali, and various prompting strategies on large language models.
\\
The key contributions of the paper are summarised as follows:
\\
\begin{itemize}

    \item   A manually annotated dataset of 4,060 Bengali comments about public versus private university debates, where there was high inter-annotator agreement (Fleiss's Kappa = 0.89)\cite{stance_detection_repo}. The dataset and all source code are publicly available on GitHub at \url{https://github.com/SafaruzzamanShovo/Educational-Ideological-Stance-Detection}.
    
    \item A comparative analysis of multiple models for stance detection in Bengali educational discourse, including task-specific fine-tuning of multilingual and Bengali language models.
    
    \item Benchmark results establishing strong baselines for stance detection on this dataset, covering classical ML, BiLSTM, BanglaBERT+XGBoost, and LLMs.
\\ 
    
\end{itemize}

The rest of the paper is organized as follows: Section II reviews related work. Section III covers the dataset, preprocessing, augmentation, models, and experimental setup. Section IV presents results and discussion. Section V concludes and outlines future work.

\section{Related Work}
In our context, ideological stance detection is the task of determining the stance of a comment whether a comment supports public universities, supports private universities, or is neutral. Initial solutions were based on classical machine learning. More recently, neural networks and pre-trained language models have improved the situation. Recently, large language models (LLMs) with prompt engineering have been able to capture the nuances of text that were not addressed by older techniques. In this section, we summarize the related works.

\subsection{Educational Discourse and Polarization in Social Media}

The use of surveys and interviews has been the traditional method of research for public versus private university preferences\cite{khondkar2024empirical}\cite{alam2024education}. In Bangladesh, research has revealed that public universities are still the first option for most students because they are cheaper and are believed to be more prestigious, and private universities are gaining popularity among urban students. 

Families who are educated in English and prefer to send their children abroad to experience new places and facilities. But these studies are usually based on closed-ended questions that do not reflect the subtleties of real-world debates.

Complementary is the approach of analysing social media. Facebook conversations about university rankings, comments on education policy and threads on career prospects have been studied. Most of this work, however, is in the context of the West and English language. The special linguistic and cultural characteristics of the educational discourse of Bengali language are not well explored.

\subsection{Stance Detection in Social Media Polarization}

Stance detection is used to identify the opinion of a text author towards a specific target, and it is crucial in various contexts like social media, news articles, and online reviews\cite{liewdiscoursedash}. The task has become more than just an analysis of court debates, and is now used in social media applications, especially when analyzing political, social, and health issues\cite{liu2022politics}. The recent development of large language models (LLMs) has enabled new capabilities in understanding context and cross-domain generalization\cite{gyawali2025evaluating}. NLP has also been used to identify attitudes from text and to apply network models to generate belief systems, and to investigate the evolution of opinion polarization.

There have been several studies that have compared various modeling techniques for stance detection. A comparison of neural networks\cite{wu2018development}, pre-trained language models, and LLMs using a Reddit dataset on the Israel-Palestine conflict\cite{ali2026social} revealed that LLMs with prompt engineering performed better than the previous methods. In another study, seven transformer-based models were tested on a political text classification task in Bengali language with accuracy of 91\% for RoBERTa\cite{liu2019roberta}. Several transformer-based models have also been evaluated for Bengali text classification tasks, demonstrating the effectiveness of transformer architectures in low-resource Bengali NLP \cite{gillioz2020overview}.

Most of the current research, however, is in English or other high resource languages. In languages with a limited number of resources such as Bangla, annotated datasets and computational studies are still limited. Recently, BanglaBias\cite{lia2025read}, the first benchmark dataset for political stance detection\cite{liu2022politics} in Bangla news article, was introduced where 200 politically relevant news articles were classified as government-leaning, government-critique, and neutral. The results of 28 LLMs demonstrated good performance in the critique detection task and significant difficulties in the neutral articles task. While our study builds on this tradition, it is in a new field, namely educational discourse on public vs. private universities in Bangladesh, and offers the first benchmark data set and extensive model comparison for this topic in Bengali.

\subsection{Research Gap and Motivation}

\begin{table}[t]
\centering
\caption{Comparison of Related Stance Detection Datasets}
\label{tab:dataset_comparison}

\scriptsize
\renewcommand{\arraystretch}{1.15}
\setlength{\tabcolsep}{3pt}

\begin{tabular}{|p{2.3cm}|p{1.7cm}|p{0.9cm}|p{2.1cm}|}
\hline

\textbf{Dataset} &
\textbf{Conflict/Topic} &
\textbf{Size} &
\textbf{Stance Classes}
\\
\hline

Imtiaz et al. (2022)\cite{imtiaz2022taking}
&
Israel--Palestine
&
493K
&
3 Classes
\\
\hline

Hernandez et al. (2022)\cite{hernandes2024llms}
&
Political Bias
&
5.8K
&
Multi-level
\\
\hline

Bhatia et al. (2018)\cite{bhatia-p-2018-topic}
&
U.S. Ideology
&
1.5K
&
2 Classes
\\
\hline

\textbf{Ours}\cite{stance_detection_repo}
&
\textbf{University Debate}
&
\textbf{4,060}
&
\textbf{Pro-Public, Pro-Private, Neutral}
\\
\hline

\end{tabular}

\end{table}

Table \ref{tab:dataset_comparison} summarizes some of existing stance detection datasets.
 Several gaps motivate our work. First, educational discourse as a domain for stance detection has received minimal attention compared to political polarization\cite{ali2026social}. Second, Bengali-language stance detection\cite{liu2022politics} lacks benchmark datasets and systematic model comparisons. Third, existing studies on university preferences in Bangladesh use survey methods that cannot capture the rhetorical strategies and ideological arguments that emerge organically in social media discussions. Our work addresses all three gaps.

 \begin{table}[ht]
\centering
\caption{Class Distribution Comparison Original and Augmented + Original Entries (After Text Cleaning)}
\label{tab:class_distribution}

\renewcommand{\arraystretch}{1.15}

\begin{tabular}{|c|c|c|}
\hline
\textbf{Label Name} & \textbf{Before Augmentation} & \textbf{After Augmentation} \\
\hline
Neutral & 1617 & 1617 \\
\hline
Pro-Public & 1050 & 1239 \\
\hline
Pro-Private & 974 & 1204 \\
\hline
\end{tabular}

\end{table}

\begin{table}[ht]
\centering
\caption{Layers of the BiLSTM}
\label{tab:bilstm_layers}

\renewcommand{\arraystretch}{1.15}

\begin{tabular}{|c|c|c|}
\hline
\textbf{Layer} & \textbf{Type} & \textbf{Output Shape} \\
\hline
Embedding &
Embedding Layer &
(None, 100, 64) \\
\hline

Bidirectional LSTM &
LSTM Layer (BiLSTM) &
(None, 64) \\
\hline

Dropout &
Regularization &
(None, 64) \\
\hline

Dense &
Fully Connected (ReLU) &
(None, 16) \\
\hline

Dropout &
Regularization &
(None, 16) \\
\hline

Dense &
Fully Connected (Softmax) &
(None, 3) \\
\hline

\end{tabular}
\end{table}

\begin{figure*}[t]
    \centering
    \includegraphics[width=\textwidth]{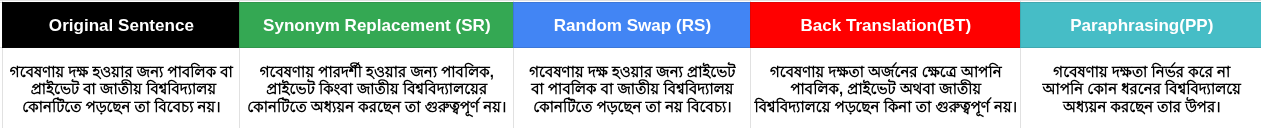}
    \caption{Example of the Augmentation Techniques from the BDA Framework}
    \label{fig:augmentation}
\end{figure*}

\section{Methodology}
\subsection{Dataset}

The comments were gathered from the public and private universities debate in Facebook and YouTube. The sources were university professors, educational groups, edtech groups and educational podcasts. The comments were filtered by keywords to keep comments related to public universities, private universities, educational quality, admission systems, tuition fees, and institutional comparison.

Finally, there were 4,060 comments in the final set after the removal of duplicates and irrelevant comments. These were manually labeled as Pro-Public (in favor of public universities), Pro-Private (in favor of private universities), and Neutral (no clear preference). Three annotators independently labeled each comment.

The final label for each comment was decided by majority voting. The label distribution showed a variety of views of higher education. Fleiss's Kappa was used to assess the reliability of the annotations which was found to be 0.89 – a high level of inter‑annotator agreement. The annotated data was then pre-processed, trained and evaluated the model.

\subsection{Dataset Preprocessing}

The dataset\cite{stance_detection_repo} has been preprocessed with bnlp-toolkit and some custom text cleaning functions to remove English letters, stopwords and emoji\cite{sarker2021bnlpnaturallanguageprocessing}. All labels, except for “Neutral”, were then augmented with data, resulting in a total of 561 augmented labels. The augmentation technique is explained in Section III‑C.

Second text cleaning step: This process eliminates the empty strings and the duplicates in the combined (original plus augmented) dataset. This step eliminated empty strings and duplicates. Note that these duplicates and empty strings have been added only after the first cleaning: When the emojis and stopwords were removed from the Bangla text, some comments became identical that were previously different due to the different stopwords or emoji's attached to the same Bangla text.

There were 79 duplicates and 63 empty strings in the augmented set. Thus, the size of the enlarged set was 419 and the size of the original set was 3,641. Out of the 419 augmented entries, 189 were labeled Pro-Public and the remaining 230 were Pro-Private. The class distribution of the combined dataset (original + augmented, totalling 4,060 comments) after both cleaning iterations is shown in Table \ref{tab:class_distribution}.

\subsection{Data Augmentation Techniques}

To solve the class imbalance problem as mentioned in section III-A, the BDA (Bangla Text Data Augmentation) framework was used as it had outstanding performance when it in the context to augmenting Bangla text \cite{tariquzzaman2024bda}. For augmentation, The following labels or classes were targeted - “Pro-Public” and
“Pro-Private” as those had the least frequencies. For each of
the sentences, the following steps were used -

\begin{itemize}
    \item \textbf{Synonym Replacement (SR):} Randomly pick up a random number of non-stop words in the sentence and substitute them with synonyms from the most similar alternatives. It's performed using a pre-trained FastText model.
    
    \item \textbf{Random Swap (RS):}  Randomly selecting two words from the sentence and swap their positions.
    
    \item \textbf{Back Translation:} Translate the text to the intermediate language, English and then translate it back to Bangla. This was done with BanglaT5. \cite{bdcc9050117}.
    
    \item \textbf{Paraphrasing: }Using a different sentence structure to convey the same meaning. Again, BanglaT5 was used in this regard.
    
\end{itemize}

The augmentation techniques demonstrated in Figure 1. Note that we selected BanglaT5 as the paraphraser and back translator because of its robust performance in both machine translation and cross-lingual summarization  \cite{bdcc9050117}.
\vspace{-0.1cm}
\subsection{Model Architectures}

We used different Machine Learning models for classification tasks. Each model represents a way of doing things. The models we used are simple. Do not have a lot of extra steps. We wanted to see how data augmentation helps with Machine Learning techniques. So we did not try to make the models themselves better.

\subsubsection{Traditional Machine Learning Models} We employed the following machine learning models: Support Vector Machine\cite{van2018hyperparameter}, Random Forest\cite{Probst_2019} and XGBoost\cite{Putatunda_2019} which are vector based, ensemble and boosting based models, respectively. These models are chosen due to their successful application in text classification applications. The optimal values of the hyperparameters were selected from the literature. All the traditional machine learning models were subjected to 5-fold stratified cross-validation and the results reported are the average performance across the 5 folds.

\subsubsection{Neural Network Architectures}

In this study, a Bidirectional Long Short-Term Memory (BiLSTM) network was used as the representative neural network model. BiLSTM was selected since it can learn bidirectional information of the given sequence, which is essential for text classification tasks\cite{article}. Table \ref{tab:bilstm_layers} gives a summary of the architecture. The batch size was set to be variable, which allowed the model to process batches dynamically, instead of having a fixed number of input samples.

\begin{table}[t]
\caption{Results Comparison of the Models in Multi-Class Stance Classification}
\label{tab:model_comparison}
\centering
\tiny
\setlength{\tabcolsep}{2.3pt}
\begin{tabular}{|c|l|l|c|c|c|c|c|}
\hline
Type & Model & Class & Precision(\%) & Recall(\%) & F1(\%) & Overall Acc.(\%) & Macro F1(\%)\\
\hline

\multirow{3}{*}{ML}
& \multirow{3}{*}{SVM}
& Pro-Public  & 90.43 & 93.56 & 91.97 & \multirow{3}{*}{89.27} & \multirow{3}{*}{89.09}\\
&& Pro-Private & 80.90 & 90.45 & 85.41 & &\\
&& Neutral & 94.68 & 85.57 & 89.89 & &\\
\cline{2-8}

& \multirow{3}{*}{RF}
& Pro-Public  & \text{89.00} & 88.12 & 88.56 & \multirow{3}{*}{87.78} & \multirow{3}{*}{87.36}\\
&& Pro-Private & 79.59 & 87.64 & 83.42 & &\\
&& Neutral & 92.73 & 87.63 & 90.11 & &\\
\cline{2-8}

& \multirow{3}{*}{XGBoost}
& Pro-Public  & 92.31 & 89.11 & 90.68 & \multirow{3}{*}{86.74} & \multirow{3}{*}{86.56}\\
&& Pro-Private & 76.73 & 87.08 & 81.58 & &\\
&& Neutral & 90.15 & 84.88 & 87.43 & &\\
\hline

\multirow{3}{*}{DL}
& \multirow{3}{*}{BiLSTM}
& Pro-Public  & 94.43 & 90.38 & 92.36 & \multirow{3}{*}{89.74} & \multirow{3}{*}{89.71}\\
&& Pro-Private & 86.24 & 91.13 & 88.62 & &\\
&& Neutral & 90.58 & 89.38 & 89.98 & &\\
\cline{2-8}

& \multirow{3}{*}{BanglaBERT+XGB}
& Pro-Public & 94.74 & 90.86 & \text{92.76} & \multirow{3}{*}{91.81} & \multirow{3}{*}{91.70}\\
&& Pro-Private & 92.83 & 87.35 & 90.00 & &\\
&& Neutral & 87.17 & \text{94.40} & 90.64 & &\\
\hline

\multirow{3}{*}{LLM}
& \multirow{3}{*}{Kimi K2 Thinking}
& Pro-Public & 93.99 & 93.19 & 93.59 & \multirow{3}{*}{90.49} & \multirow{3}{*}{90.38}\\
&& Pro-Private & 86.07 & 94.17 & 89.95 & &\\
&& Neutral & 91.84 & 83.72 & 87.59 & &\\
\cline{2-8}

& \multirow{3}{*}{DeepSeek-V3.1}
& Pro-Public & 93.48 & 91.49 & 92.47 & \multirow{3}{*}{89.15} & \multirow{3}{*}{88.95}\\
&& Pro-Private & 84.25 & 95.96 & 89.72 & &\\
&& Neutral & 90.48 & 79.53 & 84.65 & &\\
\cline{2-8}

& \multirow{3}{*}{Qwen3-235B Thinking}
& Pro-Public & 94.96 & \text{96.17} & 95.56 & \multirow{3}{*}{\text{92.12}} & \multirow{3}{*}{\text{91.96}}\\
&& Pro-Private & \text{88.75} & \text{95.52} & \text{92.02} & &\\
&& Neutral & \text{92.82} & 84.19 & \text{88.29} & &\\
\cline{2-8}

& \multirow{3}{*}{Llama 4 Maverick}
& Pro-Public & 95.04 & 97.87 & 96.43 & \multirow{3}{*}{93.31} & \multirow{3}{*}{93.11}\\
&& Pro-Private & 88.35 & 98.65 & 93.22 & &\\
&& Neutral & 97.80 & 82.79 & 89.67 & &\\
\cline{2-8}

& \multirow{3}{*}{Claude Sonnet 4}
& Pro-Public & 88.45 & 89.71 & 89.08 & \multirow{3}{*}{88.49} & \multirow{3}{*}{88.49}\\
&& Pro-Private & 86.38 & 91.17 & 88.71 & &\\
&& Neutral & 89.91 & 85.53 & 87.67 & &\\

\hline
\end{tabular}
\end{table}

\subsubsection{Transformers Architecture}
In this study, BanglaBERT is used as the transformer model. BanglaBERT is a pre-trained BERT\cite{devlin2019bertpretrainingdeepbidirectional} model trained on the union of the Common Crawl and Wikipedia Dump in Bengali language. It is similar to BERT, with 12 layers, 768 hidden units and 12 attention heads, for a total of 110M parameters. BanglaBERT surpasses the current models mBERT and Bengali Electra in the hate speech detection task of the Bengali Classification Benchmark with an accuracy of 71.83\% \cite{Sagor_2020}. This benchmark is consistent with the study's premise and BanglaBERT was chosen. As a base model, BanglaBERT is not specialized for classification, which requires a classification layer. Here, an XGBoost classifier was used. The last four hidden layers were then mean pooled and the embeddings were passed to XGBoost after the forward pass. This technique is known as boost pooling, which takes advantage of the last layers weights to represent semantic relationships between tokens. The hyperparameters of the XGBoost model were set to the same values used in the standalone model to evaluate the effect of the pooling technique.

\subsubsection{Large Language Models (LLMs)}

The five large language models used in this study are instruction‑tuned: Claude Sonnet 4 \cite{anthropic2025claude4}, DeepSeek-V3.1\cite{deepseekv3}, Qwen3-235B Thinking \cite{qwen3}, Llama 4 Maverick \cite{meta2025llama4}, and Kimi K2 Thinking\cite{kimi2026}. These models represent recent reasoning-oriented LLMs from different organizations, covering diverse architectures, training objectives, and pre-training corpora, including both multilingual and English-dominant data.
\\
No fine‑tuning or few‑shot examples were used. Each model received a simple prompt to classify a comment as pro\_public, pro\_private, or neutral. The prompt was kept short to test the model's raw zero‑shot ability.

\subsection{Experimental Setup}

The experimental setup began with data preprocessing and ended with model evaluation. Firstly, the noise and consistency were reduced by text cleaning. Then, data augmentation was applied to obtain more diversity and to mitigate class imbalance among the trainings. The original data was then duplicated and invalid samples were removed from both sets during a second cleaning prior to combining the two sets.

The dataset has been divided into 80\% training and 20\% testing. The training set was employed in constructing the model, and the test set was kept aside for final tests.

In traditional machine learning methods, the textual data was vectorized using TF‑IDF and then classified using SVM, Random Forest and XGBoost\cite{van2018hyperparameter}. At the same time, a BiLSTM\cite{article} network was directly trained on the processed text to learn the context and generate classifications.

BanglaBERT\cite{Sagor_2020} created contextual representations and the pooled features were inputted into an XGBoost\cite{Putatunda_2019} classifier for transformer-based modeling. Besides, multiple LLMs (KImi K2 Thinking\cite{kimi2026}, Qwen3-235B Thinking\cite{qwen3}, DeepSeek-V3.1\cite{deepseekv3}, Llama 4 Maverick\cite{meta2025llama4} and Claude Sonnet 4) were tested on the test set without any task specific training.

Finally, for a uniform and complete evaluation, all models were evaluated using accuracy, precision, recall, and macro F1‑score. The general experimental scheme is shown in Fig. 2.

\begin{figure*}[t]
    \centering
    \includegraphics[width=\textwidth]{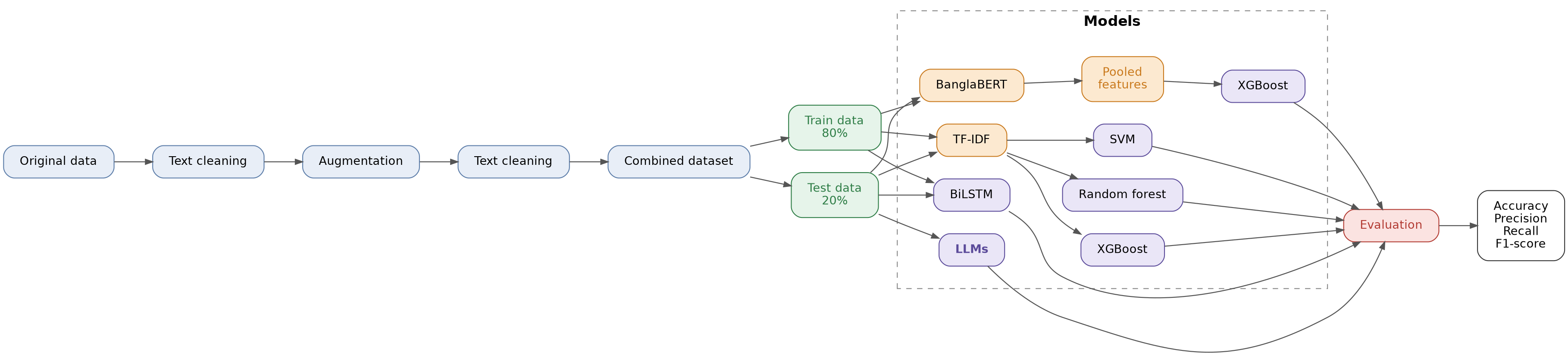}
    \caption{Overview of the Experimental Setup}
    \label{fig:exp}
\end{figure*}
\vspace{-0.3cm}
\section{RESULTS and DISCUSSIONS}

\subsection{Stance Classification and Polarization Findings}

The results are presented in table \ref{tab:model_comparison}. The SVM model performed best in the classical models, with an overall accuracy of 89.27\% and macro F1 score of 89.09\% and per-class F1 scores of 91.97\% (Pro-Public), 85.41\% (Pro-Private), and 89.89\% (Neutral). Random Forest (macro F1 87.36, accuracy 87.78\%) and XGBoost (macro F1 86.56, accuracy 86.74\%) trailed behind. Therefore, a well-trained SVM on top of TF-IDF is very effective for this task, and does not require any deep learning.

BiLSTM attained the highest macro F1 score of 89.71 and overall accuracy of 89.74\%, while the F1 scores of Pro-Public, Pro-Private and Neutral were 92.36, 88.62 and 89.98 respectively, which were fairly balanced across the classes. Transformer-based representations performed quite well on Bengali text, with BanglaBERT+XGB having macro F1 91.70 and accuracy 91.81\%, and per-class F1 of 92.76 (Pro-Public), 90.00 (Pro-Private), and 90.64 (Neutral).

Llama 4 Maverick was the best performer of the five zero-shot LLMs, with macro F1 of 93.11 and overall accuracy of 93.31\%, and per-class F1 of 96.43 (Pro-Public), 93.22 (Pro-Private), and 89.67 (Neutral); the Pro-Public recall reached 97.87\%. Qwen3-235B Thinking followed closely with macro F1 91.96 (accuracy 92.12\%), then Kimi K2 Thinking (macro F1 90.38, accuracy 90.49\%), DeepSeek-V3.1 (macro F1 88.95, accuracy 89.15\%), and Claude Sonnet 4 (macro F1 88.49, accuracy 88.49

Notably, BanglaBERT+XGB (macro F1 91.70) is very close to — and in fact slightly above — the best zero-shot LLM, Llama 4 Maverick (93.11 is higher) and outperforms most of the other LLMs despite being a much smaller, fine-tuned model. This is an impressive result because the LLM was not task-trained, indicating that these can be a good choice in low-resource labeling cases.

The Neutral class was consistently the more difficult class across all models, with F1 scores typically lower than the two pro classes (e.g., BiLSTM: 89.98 vs. 92.36/88.62; Claude Sonnet 4: 87.67 vs. 89.08/88.71). This is because neutral comments are often ambiguous and under-expressed.
\vspace{-0.3cm}
\subsection{Discussion}
\subsubsection{Overview of the Augmented Dataset}
 The original 3,641 comments were paraphrased and back-translated using BanglaT5\cite{tariquzzaman2024bda} to increase the number of comments, which were applied only to the Pro-Public and Pro-Private classes, but not the neutral class. This resulted in 561 new comments, which was reduced to 419 when any duplicate or empty strings were removed, leaving 4,060 comments for training. The class distributions before and after augmentation are given in Table II.

\subsubsection{Data Augmentation Process Critique} The augmentation process is based on BDA which produces paraphrased and back-translated samples based on BanglaT5 checkpoints \cite{tariquzzaman2024bda}. As explained in Section III-B, 79 samples were found to be duplicates and discarded from the 561 samples generated, which is about 14\%. This represents a drawback to the augmentation approach because a few of the augmented samples are still quite similar to the originals, with only a few lexical changes.

While SacreBLEU-based filtering can help to remove semantically unrelated sentences generated by the model, it does not set an upper bound on the similarity threshold to ensure that the generated sentences are not too similar. Thus, the augmentation process may be less effective. In future, maximum similarity filtering (discarding generated samples that have similarity score greater than a pre-defined BLEU threshold) and word level replacement techniques with Bengali WordNet resources could be integrated to enhance lexical diversity.

\subsubsection{Why LLMs (Zero-Shot) Perform Better}
Interestingly, without fine-tuning or few-shot examples, the zero-shot prompting approach achieves a higher performance than BanglaBERT+XGBoost (Macro F1: 0.917) and other ML/DL models with a Macro F1 score of 0.931 for the Llama 4 Maverick model.

This improvement could be due to the size of the pre-training data and its exposure to a wide variety of multilingual and code-mixed data, which allows the model to learn contextual and pragmatic information like sarcasm and implied sentiment. Fine-tuned models trained on our small set of augmented data, on the other hand, might lack the generalizability.

This observation, however, is a plausible explanation, since the effects of pre-training scale, multilingual exposure, prompt design and statistical significance have not been independently controlled and evaluated.

\subsubsection{Threats to Validity} Construct validity might be impacted by the subjective nature of stance annotation. While the agreement for the annotation of the neutral class was high (Fleiss's Kappa = 0.89), there were still cases of ambiguity that could potentially be interpreted differently by humans and LLMs.

The augmentation process could impact the internal validity because of the fact that about 14\% of the generated samples were duplicates, which might raise the danger of overfitting. Furthermore, the lack of statistical significance testing, augmentation ablation studies and quantitative error analysis reduces the quality of the comparative evaluation.

The data set is relatively small as it only includes Bengali code-mixed comments from Facebook and YouTube. So the results may not be applicable to other platforms, languages or domains. Although these limitations exist, the use of a common augmented dataset and the same evaluation conditions allows for a fair comparison of the evaluated models.

\section{Conclusion and Future Work}

We introduced a new manually annotated Bengali dataset consisting of 4,060 comments to investigate the polarization of social media on the public versus private university issue in Bangladesh. By extension, with the help of BanglaT5, the Llama 4 Maverick model scored the highest macro F1 of 0.931, outperforming all the supervised baselines. But, there were 14\% duplicates (79 out of 561) in the augmentation, which revealed the problems in the existing Bangla paraphrasing tools. Next, we will fine-tuned on Bengali, and an upper similarity filter to remove near-duplicates for BanglaT5. The ultimate aim is a real-world, open source stance detector to facilitate the study of polarization in social media without requiring costly commercial application programming interfaces (APIs).

\balance
\bibliographystyle{IEEEtran}
\vspace{-0.3cm}
\bibliography{references}

\end{document}